\documentclass[10pt,twocolumn,letterpaper]{article}

\usepackage{iccv}
\usepackage{times}
\usepackage{epsfig}
\usepackage{graphicx}
\usepackage{amsmath}
\usepackage{amssymb}

\usepackage{multirow}
\usepackage{caption}
\usepackage{subcaption}

\usepackage[pagebackref=true,breaklinks=true,letterpaper=true,colorlinks,bookmarks=false]{hyperref}

\iccvfinalcopy %

\def\iccvPaperID{****} %

\ificcvfinal\fi

\newcommand{\bcdot}{\boldsymbol{\cdot}}

\begin{document}

\title{Contrast Learning Visual Attention for Multi Label Classification}

\author{Son D.Dao \qquad Ethan Zhao \qquad Dinh Phung \qquad Jianfei Cai\\
Faculty of Information Technology\\
Monash University\\
}

\maketitle
\ificcvfinal\thispagestyle{empty}\fi

\begin{abstract}
    Recently, as an effective way of learning latent representations, contrastive learning has been increasingly popular and successful in various domains. The success of constrastive learning in single-label classifications motivates us to leverage this learning framework to enhance distinctiveness for better performance in multi-label image classification. In this paper, we show that a direct application of contrastive learning can hardly improve in multi-label cases. Accordingly, we propose a novel framework for multi-label classification with contrastive learning in a fully supervised setting, which learns multiple representations of an image under the context of different labels. This facilities a simple yet intuitive adaption of contrastive learning into our model to boost its performance in multi-label image classification.
    Extensive experiments on two %
    benchmark datasets show that the proposed framework achieves state-of-the-art performance in the comparison with the advanced methods in multi-label classification.
\end{abstract}

\section{Introduction}
Multi-label image classification is a fundamental and practical computer vision task, where the goal is to predict a set of labels (e.g., objects or attributes) associated with an input image. It is an essential component in many application such as recommendation systems \cite{jain2016extreme, yang2015pinterest}, medical image diagnosis \cite{ge2018chest}, and human attribute recognition \cite{li2016human}. Compared to single-label cases, multi-label classification is usually more complex and challenging.

Recently, contrastive learning (CL)~\cite{chen2020simple, he2020momentum} is effective in learning latent representations for supervised tasks.
In general, CL aims to pull together an anchor and a similar (or positive) sample in embedding space and push apart the anchor from many dissimilar (or negative) samples. Therefore, the choice of the positive and negative samples of an anchor is a key to achieving good performance with CL.
In self-supervised CL~\cite{chen2020simple}, the positive sample is defined as those augmented from the same image with the anchor, while the negative samples are all the other images in the minibatch.
More recently, supervised CL~\cite{khosla2020supervised} has been proposed, where all the images with the same label as the anchor are considered as the positive samples and vice versa for the negative ones. Supervised CL has shown improvements in single-label image classifications than the self-supervised counterpart.
With the above successful examples, CL has drawn significant research attention and has been applied in other tasks including image segmentation \cite{wang2021exploring}, adversarial training \cite{kim2020adversarial} or text to image learning \cite{radford2021learning}. 

Given the appealing properties and promising results of CL in single-label classification, it is natural to adapt it into multi-label cases to boost performance. 
However, this adaptation is non-trivial.
In single-label cases, an image usually contains one salient object, thus, the label of the object can also be viewed as the unique label of the image. Therefore, it is reasonable to use one image-level representation of an image and to push the representation of the anchor close to its positive samples (e.g, augmentations of the anchor or images with the same label as the anchor), as done in self-supervised and supervised CL.
However, with a single image-level representation for an image, it is hard to define the positive or negative samples for an anchor image by its multiple labels in the multi-label classification. For example, it would not be reasonable to assume that the image-level representations of images containing apples must always be close to each other, as apple is just one of many objects that those images contain and an apple may only take a small area of an image. As a result, this setting hinders the application of existing CL methods to multi-label classification.

To bridge this gap, we in this paper propose a novel end-to-end framework for multi-label image classification that leverages the contrastive learning principle, termed \textbf{MulCon} or \emph{Multi-label Classification with Contrastive Loss}.
Instead of using image-level representations as in previous CL methods, we introduce a new module that learns multiple label-level representations of an image, which are generated with the attention from the globally class-specific embeddings to the image features learned by a convolutional neural network (CNN). 
Each label-level embedding of an image corresponds to its representation in the context of a specific label, thus, it is associated with a single label.
If we view an label-level embedding as the anchor in our proposed CL framework, it is straightforward to define the positive samples of an anchor, which are the label-level embeddings of other images with the same label and vice versa for the negative samples.
With these definitions, the supervised CL loss can be applied.
Our framework is intuitive in the multi-label setting. For example, instead of sharing an image-level representation with other objects, the apple object of an image has its own embedding and the embeddings of the apples of all the images in a minibatch are pushed close to each other.
In this way, the CL loss can enforce the coherence and consistency of the label-level representations of images, which further provides more discriminative power of the prediction procedure based on these representations.

The main contributions can be summarized as follows: 
\begin{enumerate}
    \item We formalize multi-label classification for natural images as a contrastive learning problem, where we enhance image relationship by decomposing image into several components and enforcing the correlation of the same label components among images. To our knowledge, this is the first work applying contrastive learning in multi-label classification for natural images.
    \item We design a novel end-to-end multi-label classification framework to decompose image into several semantic components. A two-step training procedure that combines the contrastive loss with the birnary cross-entropy loss is introduced to improve the quality of these semantic components. 
    \item We conduct extensive experiments on large-scale benchmark datasets including COCO~\cite{lin2014microsoft} and NUS-WIDE~\cite{chua2009nus}, showing that the proposed framework achieves the state-of-the-art performance in multi-label image classification.
\end{enumerate}

\section{Proposed Method} 
In this section, we first discuss about the background knowledge and relevant notations and then elaborate on the details of the proposed framework, MulCon.

\subsection{Background and Notations}
\noindent\textbf{Multi-label Classification.}
Following the standard setting of the multi-label image classification, we denote a batch of input images by $X \in \mathbb{R}^{N \times W \times H \times 3}$, where $N$ is the batch size, $H$ and $W$ are the height and width of the images. Each image $x_i \in X$ is associated with multiple labels selected from a set of $L$ labels in total, which are denoted by a multi-hot binary vector $y_i \in \{0,1\}^L$. For an active label $j$ of $x_i$, $y_{ij} = 1$ and vice versa.
Our task is to build an end-to-end model that takes $x_i$ to predict its labels $y_i$.

\noindent\textbf{Attention.}
The Attention mechanism \cite{luong2015effective, xu2015show} 
has been widely used in various areas of computer vision and natural language processing, which enhances the important parts of the data of interest and fades out the rest. 
Assume that $n_q$ query vectors of size $d_q$ denoted as $Q \in \mathbb{R}^{n_q \times d_q}$, and $n_v$ key-value pairs denoted as $K \in \mathbb{R}^{n_v \times d_q}$, $V \in \mathbb{R}^{n_v \times d_v}$. The attention function maps the query vectors $Q$ to outputs using the key-value pairs as follows:
\begin{align}
    \text{Att}(Q, K, V) = \omega (QK^T)V
\end{align}
where the dot product $(QK^T) \in \mathbb{R}^{n_q \times n_v}$ and $\omega(\bcdot)$ is softmax function. The dot product returns the similarity of each query and key value. The output $\omega (QK^T)V \in \mathbb{R}^{n_q \times d_v}$ is the weighted sum over $V$, where larger weight corresponds to larger similarity between query and key.

A powerful extension to the above (single-) attention mechanism is the multi-head attention introduced in \cite{vaswani2017attention}, which allows the model to jointly attend to information from different representation subspaces at different positions.
Instead of computing a single attention function, this method first projects $Q, K, V$ onto $h$ different vectors, respectively. An attention function $Att(\bcdot)$ is applied individually to these $h$ projections. The output is a linear transformation of the concatenation of all attention outputs:
\begin{align}
    \label{eq:mha}
    &\text{MultiAtt}(Q,K,V) =  \text{concat}(O_1,O_2,\cdots,O_h)W^o,\nonumber\\
    &O_{i'} =  \text{Att}(QW^q_{i'}, KW^k_{i'}, VW^v_{i'}) \text{~for~} i' \in {1,\cdots,h},
\end{align}
where $W^o, W^q_{i'}, W^k_{i'}, W^v_{i'}$ are learnable parameters of some linear layers. $QW^q_{i'} \in \mathbb{R}^{n_q \times d^h_q}$, $KW^k_{i'} \in \mathbb{R}^{n_v \times d^h_q}$, $VW^v_{i'} \in \mathbb{R}^{n_v \times d^h_v}$ are vectors projected from $Q,K,V$ respectively. $d^h_q = d_q/h$ and $d^h_v = d_v/h$.

Following the architecture of the transformer~\cite{vaswani2017attention,lee2019set}, we further define the following multi-head attention block:
\begin{align}
    &\text{MultiAttBlock}(Q,K,V) = Q' + Q'W^{q'},\\
    &Q' = \text{concat}(QW^q_{1},\cdots, QW^q_{h}) + \text{MultiAtt}(Q, K, V) \nonumber
\end{align} where $W^{q'} \in \mathbb{R}^{d_q \times d_q}$ is a learnable linear layer. %

\noindent\textbf{Contrastive Learning.}
Contrastive learning (CL) has been an increasingly popular
and effective representation learning approach \cite{chen2020simple,he2020momentum,khosla2020supervised}.
As the first proposed CL approach, self-supervised CL~\cite{chen2020simple} is proposed to learn presentations in an unsupervised manner.
Specifically, a mini-batch is constructed from the original input images and their augmented versions. Given a minibatch of $2N$ instances $I = \{1...2N\}$ and an anchor instance $i \in I$, the augmented version of $i$, denoted as $i_a \in I$ is considered the positive sample, and the other $2(N-1)$ instances within the mini-batch are considered negative examples. The loss function of self-supervised CL is defined as follows:
\begin{align}
    \mathcal{L}_{self} = \sum\limits_{i \in I} L_i = -\sum\limits_{i \in I} \log \frac{\exp(z_i \bcdot z_{i_a}/\tau)}{\sum\limits_{z_a \in A(i)} \exp(z_i \bcdot z_a/\tau)}
\end{align}
where $z_i$ is the image embedding, $z_i \bcdot z_{i_a}$ denotes the inner dot product between two embeddings, $\tau \in \mathbb{R}^+$ is a scalar temperature parameter, and $A(i) = I \backslash z_i$.
It can be seen that self-supervised CL pushes the embeddings of the samples augmented from the same image close to each other.

More recently, the supervised CL \cite{khosla2020supervised} adapts CL into the supervised settings, which utilizes the label information to select positive and negative samples. The supervised CL loss is:
\begin{align}
    \label{eq:scl}
    \mathcal{L}_{sup} = \sum\limits_{i \in I}\frac{-1}{\vert P(i) \vert}\sum\limits_{p \in P(i)} \log \frac{\exp(z_i \bcdot z_p/\tau)}{\sum\limits_{z_a \in A(i)} exp(z_i \bcdot z_a/\tau)}
\end{align}
where $P(i)$ is the positive set of $z_i$, which contains all the other samples of the same label with $z_i$ in the minibatch and $\vert P(i) \vert$ is its cardinality.
For supervised CL, the embeddings of the samples with the same labels are pushed close to each other, which achieves better performance in classification tasks. 

\begin{figure}[t]
\begin{center}
\includegraphics[width=\linewidth]{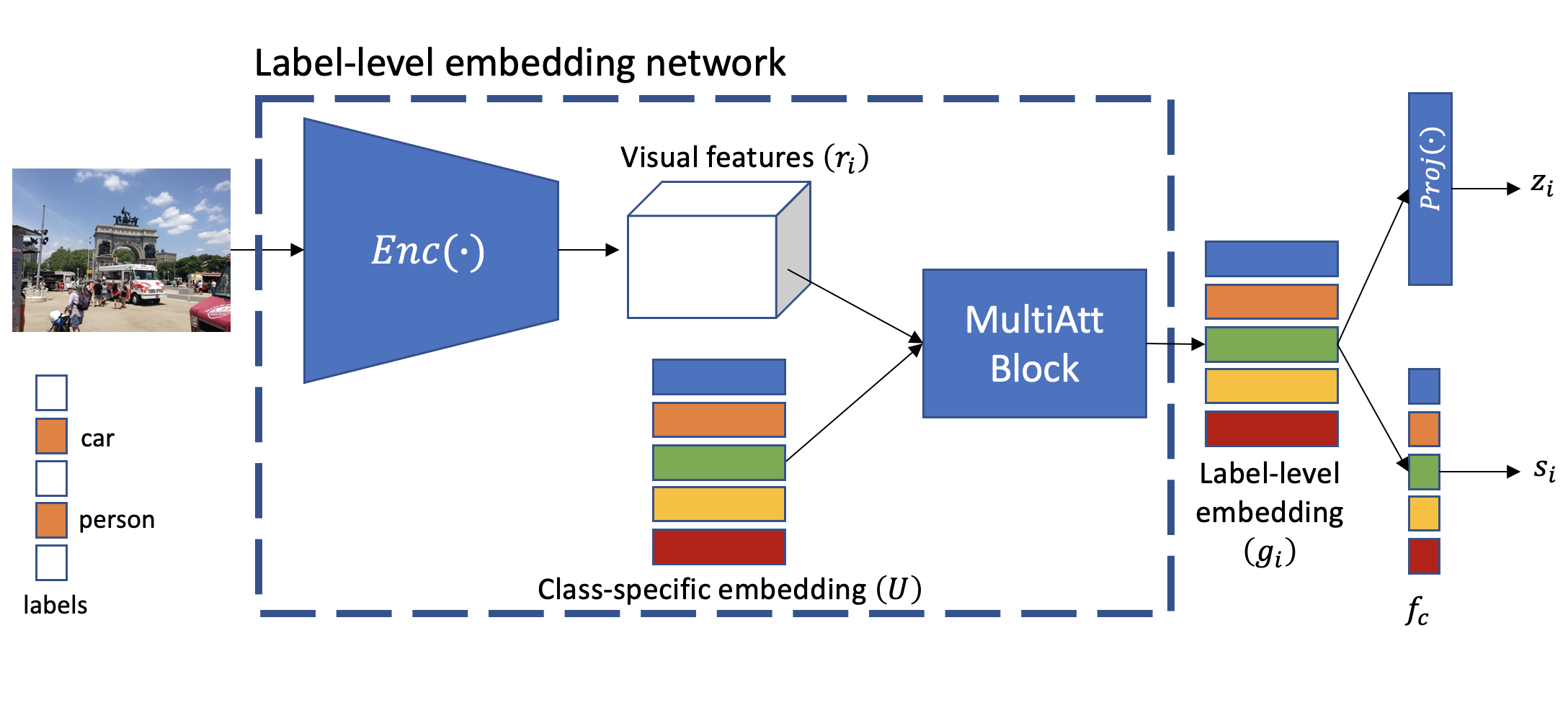}
\end{center}
\caption{Our \textbf{MulCon} framework consists of three main components: label-level embedding network that uses a multi-attention block with an Encoder ($\text{Enc}(\bcdot)$) for extracting label-level embeddings from input images, a set of independent classifiers $f_c$ for multi-label prediction, and a projector ($\text{Proj}(\bcdot)$) to map label-level embeddings to a latent space for contrastive learning.}
\label{fig:overall_framework}
\end{figure}

\begin{figure}%
\begin{center}
\includegraphics[width=\linewidth]{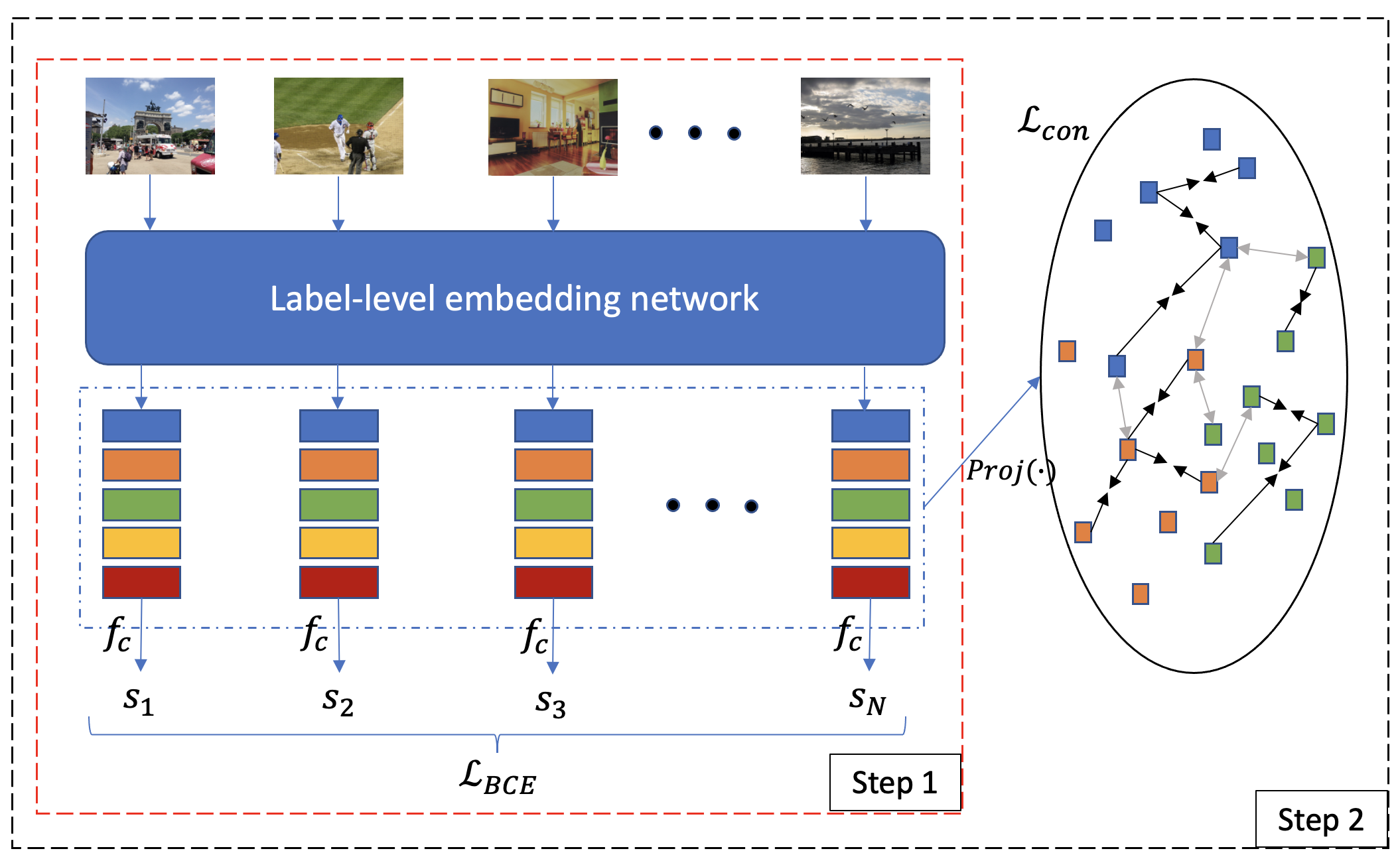}
\end{center}
\caption{\textbf{Mulcon} has two steps during training:  pretraining and contrastive finetuning. The first step is to train the label-level embedding network with binary cross-entropy loss ($\mathcal{L}_{BCE}$) to effectively decompose an input image into several semantic components so that the first component  corresponds to the first label, etc. The second step is to finetune the previously trained network with contrastive loss ($\mathcal{L}_{con}$) and $\mathcal{L}_{BCE}$ to improve the quality of label-level embedding.}
\label{fig:overall}
\vspace{-0.1in}
\end{figure}

\subsection{Proposed MulCon Framework}

Unlike previous CL methods for single-label classification that use image-level representations, we propose to learn multiple label-level representations for each image,
which facilitates the application of CL in multi-label classification.
We introduce MulCon, which consists of three main neural network modules: the label-level embedding network, the contrastive learning projection network, and the classification networks as shown in Figure \ref{fig:overall_framework}.

\noindent\textbf{Label-Level Embedding Network.}
As the key module of our framework, the label-level embedding network takes an image $x_i$ as input and outputs its label-level representations, denoted by $g_i \in \mathbb{R}^{L \times D}$, where each row of $g_i$ corresponds to the embedding of the image under the context of a specific label. Specifically, the label-level embedding network consists of three components: \textbf{1)} We first adopt an encoder network with a CNN (e.g., ResNet \cite{he2016deep}) as the backbone model, which learns the image-level embedding: 
$r_i = \text{Enc}(x_i) \in \mathbb{R}^{C \times H \times W}$, where $C, H, W$ are the number of channels, height, and width of the CNN output. We then reshape $r_i \in \mathbb{R}^{WH \times C}$ for the next step's processing.
\textbf{2)} We introduce a set of vectors $U \in \mathbb{R}^{L \times C}$, each row of which is a class-specific the embedding. $U$ is randomly initialized with normal distributions and is a parameter to be learned during training.
\textbf{3)} To generate an image's label-level embeddings $g_i$, we propose the third component to capture the interactions between $r_i$ and $U$ based on multi-head attention: \begin{align}
    g_i = \text{MultiAttBlock}(U, r_i, r_i).
    \vspace{-0.1in}
\end{align}

Given an image $x_i$, we first learn its representation $r_i \in \mathbb{R}^{WH \times C}$ with a CNN, each row of which captures embedding of a location in the image. By the multi-head attention mechanism with the class-specific embeddings $U$ as the query, the model can learn the attention weights of a label to a specific embedding.
As the input's channel and output's channel of the attention are $C$ and $D$, respectively, we have $n_q=n_v=C$ and $d_q=n_q=D$ for Eq.~\ref{eq:mha}.
For example, if an image contains an apple, the corresponding embedding is expected to be associated with a large attention weight for the apple label.
With the multi-headed attention, a label can pay its attention to multiple objects in an image, i.e., each of the attention heads can generate attention scores for a class-specific embedding over all the image-level embeddings.
In this case, if an image consists of multiple apples, each of the apples receives a specific attention score from the apple label.
With the attention from all the labels, we can derive the label-level embeddings $g_i$.

\noindent\textbf{Contrastive Learning Projection Network.}
After obtaining $g_i \in \mathbb{R}^{L \times D}$, 
we use $g_{ij} \in \mathbb{R}^{D}$ to denote the label-level embedding of the input image $i$ under the context of a specific label $j$  ($j \in \{1, \cdots, L\}$).
Following \cite{chen2020simple, khosla2020supervised}, our framework includes a projection network $\text{Proj}(\cdot)$ that maps $g_{ij}$ to a vector in another embedding space: $z_{ij} = \text{Proj}(g_{ij}) \in \mathbb{R}^{d_z}$,
where the contrastive learning is performed. 

\noindent\textbf{Classification Network.}\label{sec:class_net}
Recall that the label-level embedding $g_{ij}$ captures the input image $i$'s feature under the context of the label $j$.
Thus, it can be used to predict whether $j$ is active in $i$.
Accordingly, we introduce a fully connected layer as a classifier $f_c^j$ to predict the probability of the label $j$ being active. Specifically, for each label $j \in L$ the prediction score is $s_{ij} = \sigma(f_c^j(g_{ij})) \in (0,1)$. We further denote $s_{i} \in (0,1)^L$.

\subsection{Learning MulCon with Contrastive Loss}\label{sec:method}
After introducing the framework, we describe the learning process of MulCon by showing the loss function first. Based on the label-level embeddings of the image $i$, i.e., $g_i$, we introduce a loss function with two terms: the classification loss and contrastive loss. The overall training process is illustrated in Figure \ref{fig:overall}.

\noindent\textbf{Classification Loss}
\label{sec:attention}
For the classification loss, given the predictive probabilities output from the classification network $s_i$ and the ground-truth multi-hot label vector $y_i$, we apply the binary cross-entropy (BCE) loss, which has been widely used for multi-label classification:
\begin{align}
    \label{eq:bce}
    \mathcal{L}_{BCE} = \sum_{j=1}^L y_{ij} \log s_{ij} + (1-y_{ij})\text{log}(1 - s_{ij})
\end{align}
It is noteworthy that other multi-label classification losses than BCE can also be used in our framework, e.g., in \cite{ben2020asymmetric, lin2017focal}.

\noindent\textbf{Contrastive Loss}
In addition to the classification loss, we introduce the contrastive loss for multi-label classification, which is one of the key contributions of this paper.
Because an image is associated with multiple labels, it is hard to directly apply CL as we analyzed before.
However, in MulCon, after learning the label-level embeddings for an image, we show that the multi-label problem can be transformed into a single-label one, where CL can be adapted in a straightforward way.

As CL works in the projected space, hereafter, we also call $z_i$ projected from $g_i$ the label-level embeddings for image $i$. Given a minibatch of $N$ images, we first forward-pass them through the label-level embedding network and contrastive learning projection network and then aggregate the label-level embeddings of all the images into set 
$Z = \{z_{ij} \in \mathbb{R}^{d_z} \vert i \in \{1, \cdots, N\}; j \in \{1, \cdots, L\}\}$. 
Similarly, we define the set of the ground-truth labels of the minibatch: $Y = \{y_{ij} \in \{0,1\} \vert i \in \{1, \cdots, N\}; j \in \{1, \cdots, L\}\}$.
If we view an image's label-level embedding $z_{ij}$ as an instance instead of the image itself, $z_{ij}$ is associated with a single ground-truth label $y_{ij}$. 
We further define $I = \{z_{ij} \in Z| y_{ij} = 1\}$ as the set that contains the label-level embeddings with active ground-truth labels and $A(i,j) = I \setminus z_{ij}$ as the set contains the embeddings in $I$ with $z_{ij}$ excluded.

In the minibatch, we now consider $z_{ij} \in I$ as the anchor, which presents the feature of image $i$ under active label $j$. With CL, we aim to push $z_{ij}$ closer to the embeddings under the same active label $j$ of other images in the minibatch, i.e., the positive set, defined as $P(i,j) = {\{z_{kj} \in A(i,j)| y_{kj} = y_{ij} = 1\}}$.
With above notations and inspired by the supervised CL, we define the contrastive loss for the anchor $z_{ij}$ as:
\begin{align}
\label{eq:contrastive_loss}
L_{con}^{ij} = \frac{-1}{|P(i,j)|}&\sum\limits_{z_p \in P(i,j)}
\log\frac{\exp(z_{ij} \bcdot z_p/\tau)}{\sum_{z_a \in A(i,j)} \exp(z_{ij} \bcdot z_a/\tau))},
\end{align}
The contrastive loss for the whole minibatch is simply the summation over all the anchors: $\mathcal{L}_{con} = \sum_{z_{ij} \in I} \mathcal{L}_{con}^{ij}$.

Together with the classification loss, we show the overall training loss of MulCon:
\begin{align}
\label{eq:final_loss}
    \mathcal{L} = \mathcal{L}_{BCE} + \gamma  \mathcal{L}_{con}.
\end{align}
where the parameter $\gamma$ controls the trade-off between the two losses.

\begin{figure}%
     \centering
     \begin{subfigure}[b]{0.22\textwidth}
         \centering
         \includegraphics[width=\textwidth]{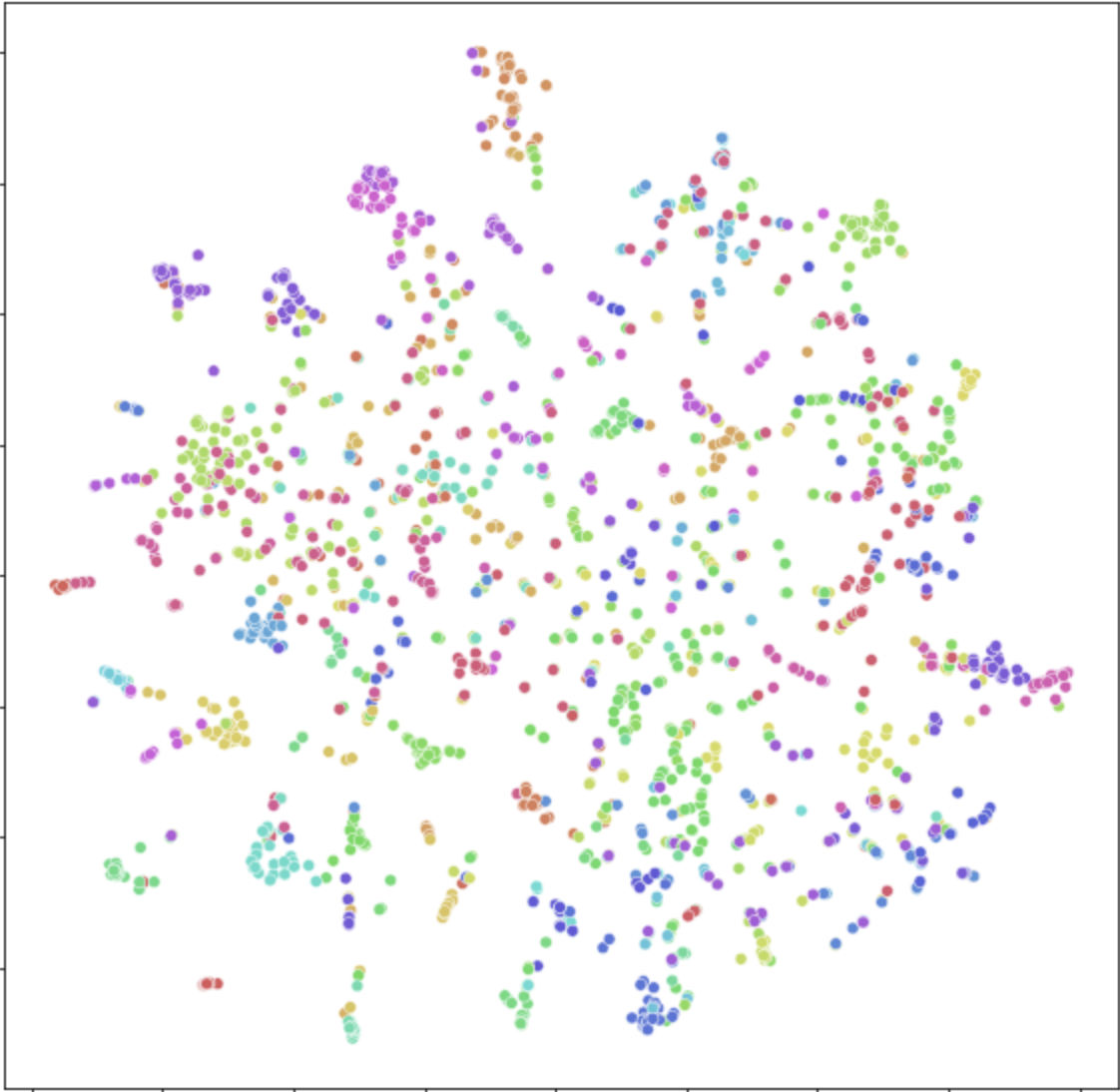}
     \end{subfigure}
     \begin{subfigure}[b]{0.22\textwidth}
         \centering
         \includegraphics[width=\textwidth]{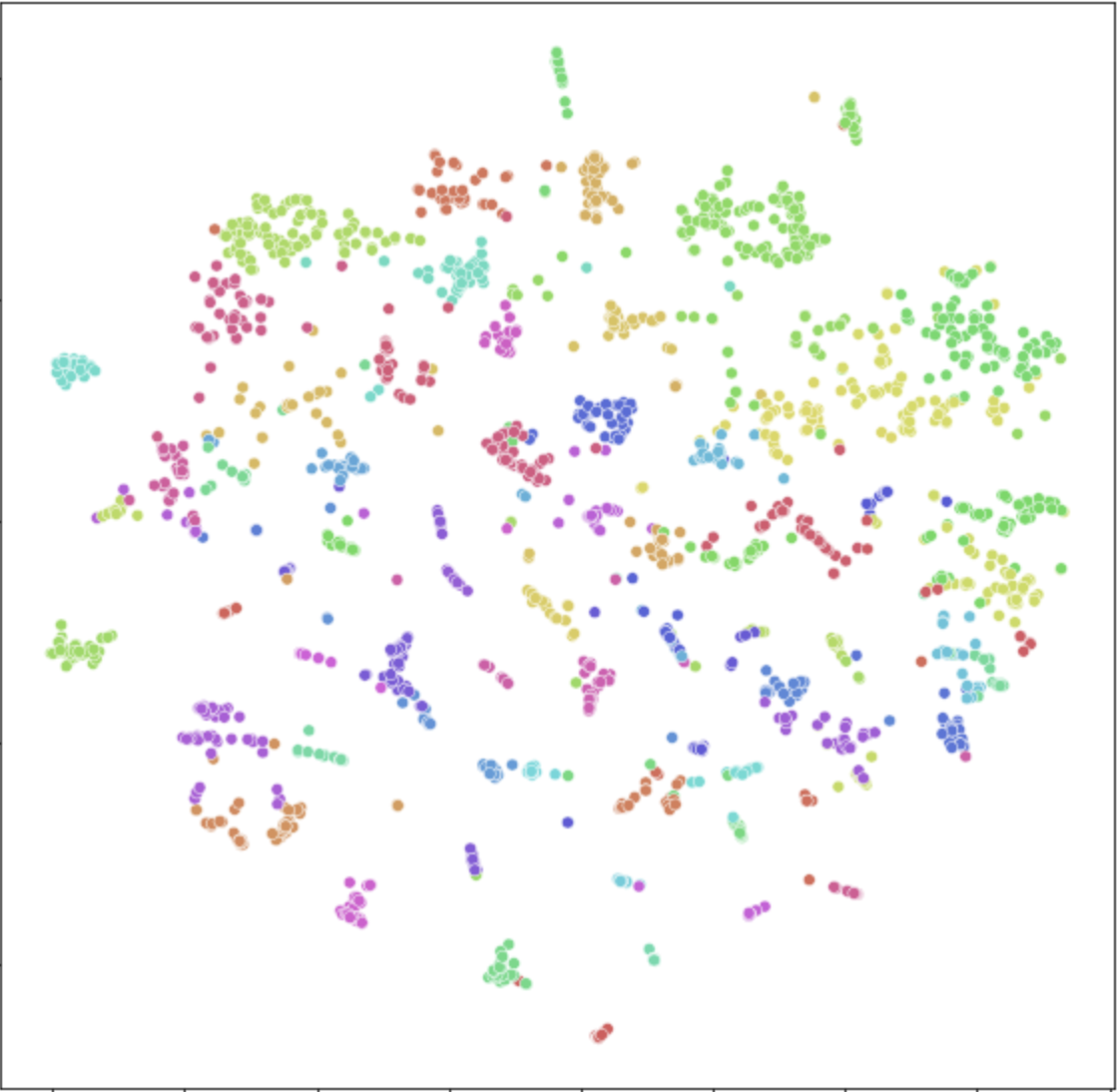}
     \end{subfigure}
        \caption{t-SNE visualization for image components trained with only $\mathcal{L}_{BCE}$ (left) and with the combination of $\mathcal{L}_{BCE}$ and $\mathcal{L}_{Con}$  (right). Each dot represents one label-level embedding under the context of a specific label and each color represents one class.}
 \label{fig:t_sne}
 \vspace{-0.1in}
\end{figure}

\noindent\textbf{Why and how does contrastive loss help?}
\label{sec:why_con_work}
Now we would like to answer why contrastive loss helps in multi-label classification, by showing that it serves as an important complementary to the classification loss.
Specifically, with the BCE classification loss applied in many multi-label classification problems, each label can be viewed to be classified independently with a specific classifier as discussed in Section~\ref{sec:class_net}. 
That is to say, each classifier focuses on the classification of a specific label and cares less about the \textit{distinctiveness} of the features of different labels. Distinctiveness in classification means that we expect features of the instances with the same label to be close to each other, which has been known as an important factor for achieving good classification accuracy.
Similar to single-label problems, the contrastive loss in MulCon is designed to enforce distinctiveness of the label-level embeddings. 
To demonstrate this, we show the t-SNE~\cite{van2008visualizing} visualization of the (active) label-level embeddings of 1,000 randomly sampled images of the COCO dataset~\cite{lin2014microsoft} in Figure~\ref{fig:t_sne}. 
These embeddings are from MulCon trained by the BCE loss (i.e., Eq.~\ref{eq:bce}) and the combined loss of BCE and CL (i.e., Eq.~\ref{eq:final_loss}), respectively.
It can be seen that compared with the loss with BCE only, the additional contrastive loss makes the label-level embeddings of the same label fall into more compact clusters, which are better separated from the clusters of other labels. This clearly shows the enforced distinctiveness of the embeddings with the contrastive loss.

However, unlike in single-label classification, being over distinct in the embedding space is not always a good thing in multi-label classification.
Specifically, label correlations are important in multi-label classification \cite{chen2019multi, you2020cross}. For example, when we see a cup in an image, the probability of a table existing below the cup in the image can be assumed to high.
In this case, we actually do want the embeddings of cups and tables are too far or too separated from each other.
Therefore, the enforced distinctiveness with the contrastive loss needs to be judiciously controlled.
In the loss of Eq.~\ref{eq:final_loss}, we can tune $\gamma$ to control distinctiveness, which might usually be time consuming.
Instead, we propose a simpler training strategy to tackle this task, which consists of two steps: \textbf{the pre-training step and the contrastive learning step}.
In the pre-training step, we train the label-level embedding network and the classification network of MulCon with the BCE loss only.
In the contrastive learning step, we then plug in the contrastive projection network with the contrastive loss but also keep the BCE loss.
In the first step, the BCE loss learns the label-level embeddings freely. After the embeddings are learned, we finetune them with the contrastive loss to enforce distinctiveness of the embeddings. We empirically find that the propose training policy works well in practice.

\section{Related Work}
\noindent\textbf{Contrast Learning} is
a state-of-the-art technique for representation learning and has received great research attention recently~\cite{bachman2019learning, chen2020simple, he2020momentum, hjelm2018learning, oord2018representation,li2020prototypical,khosla2020supervised}. In general, the works attempt to learn an embedding space where instances of the same class or augmented from the same image are pulled closer and instances from different classes or different images are pushed away. Among the variants of CL, we consider the supervised CL~\cite{khosla2020supervised} as the closest related work to ours, which proposes to leverage label information for single-label classification. As discussed before, the mechanism of supervised CL is not directly applicable to multi-label cases.
All of these works focus on single label classification problem, while we adapt the loss to our multi-label classification problem.

\noindent\textbf{Multi-Label Classification.}
has been a challenging and important problem in computer vision.
The simple and straightforward approach is to formulate it multiple independent single-label problems.
However, this approach ignores the topology structure between objects and the semantic dependencies among multiple classes, which is especially important for multi-label classification \cite{wang2016cnn}. Therefore, incorporating such semantics has been an important research direction in multi-label classification, where various approaches have been proposed.
Specifically, some prior works \cite{wang2016cnn, yang2016improving, chen2018order} explicitly capture the class correlation by a CNN-based model followed by a Recurrent Neural Network (RNN) \cite{medsker2001recurrent}. However, they usually suffer from the difficulty in optimize its parameters \cite{liu2017semantic}.
There are also approaches based on on probabilistic graphical model \cite{li2016conditional, li2014multi}, which model label dependencies in the covariance of probabilistic distributions, but they tend to have higher computational complexity in the statistical inference processes.
As label correlations can be formulated into graphs, some approaches apply Graph Convolutional Network (GCN) \cite{zhou2018graph} on these graphs to learn label presentations such as in \cite{chen2019multi,you2020cross}.
Moreover, \cite{yang2016improving} introduces semantic representations of images and exploit the image correlation via a structure graph. However, it is difficult to learn at a scale because it requires saving the whole dataset during training. \cite{huynh2020interactive} proposes an adaptive framework that combines both label correlation and image correlation to enforce the smoothness of labels and features of images on the data manifold. 
Another research line close to ours is using visual attention for multi-label classification.
For instance, \cite{chen2018recurrent, chen2018order, wang2017multi, ba2014multiple, huynh2020shared} use visual attention together with an RNN to sequentially extract salient objects from input image. 
The RNN controls the attention and helps point out the region of interest. 
Different from these methods, we employ a simple attention mechanism \cite{lee2019set} to implicitly locate all objects in image by one forward pass, so we can avoid the complexity introduced in RNN. Closest to our work is \cite{you2020cross}, where they use a simple attention mechanism to decompose input image into several semantic components and introduce label correlation to these components.
Some other works \cite{chen2019learning, ye2020attention} also introduce label dependency with visual attention to improve multi-label classification. On the other hand, we focus on image correlation with our attention mechanism and contrastive loss.

\section{Experiments}
In this section, we compare our proposed model with the state-of-the-art multi-label classification methods on two popular benchmark datasets:  MS-COCO~\cite{lin2014microsoft} and NUS-WIDE~\cite{chua2009nus}. In addition, comprehensive ablation and qualitative studies of the proposed method are also provided.

\noindent\textbf{Implementation Details.}
For all experiments, We adopt ResNet-101~\cite{he2016deep} as the backbone CNN model to extract image-level features, which is pre-trained on ImageNet~\cite{deng2009imagenet}. The projector $\text{Proj}(\bcdot)$ has two linear layers with $\text{ReLU}(\bcdot)$ activation. For the first-step training, we use Adam optimizer~\cite{kingma2014adam} and 1-cycle policy with the maximum learning rate is 2e-4. The training batch size is set to 64, and the dimension of the label-level feature, i.e., $D$, is 1024. For the second-step training, we use Stochastic Gradient Descent (SGD) optimizer with momentum 0.9 and weight decay is 1e-4. The learning rate is initialized as 0.01 and then reduced by a factor of 10 for every 20 epochs. The batch size for this step is 32. The temperature $\tau$ is 0.2, and $\gamma$ is 0.1. The number of head in multi-head attention is empirically set to 4, which gives us the best results.

Similar to the previous contrastive learning frameworks~\cite{chen2020simple,khosla2020supervised}, we also have an augmentation module to transform each training image before feeding them to the network. Specifically, for the augmentation, we first resize the input image to $448 \times 448$, and then apply random horizontal flip, and a Random Augmentation module \cite{cubuk2020randaugment}. Note that for the second step training with batch size 32, the actual batch size becomes 64 after data augmentation. 

\noindent\textbf{Evaluation metrics.}
To evaluate the classification performance, we mainly use the mean average precision (mAP) that is the most commonly-used metric for multi-label classification, following many other works~\cite{wang2016cnn, chen2019multi, you2020cross}. In addition, we also report the overall F1 score (OF1) and class F1 score (CF1) as the complementary metrics.
\subsection{Results on MS-COCO}
MS-COCO is a widely used dataset to evaluate computer vision tasks such as object detection, semantic segmentation, or image captioning, which has been also used as a benchmark dataset for multi-label image classification. 
In the standard setting for multi-label classification, it contains 122,218 images with 80 different categories, where every image is associated with 2.9 labels on the average. The dataset is divided into 82,081 images for training and 40,137 images for validation.
Here we compare our method with state-of-the-art methods such as CNN-RNN~\cite{wang2016cnn},  SRN~\cite{zhu2017learning}, s-CLs~\cite{liu2018multi}, Multi Evidence~\cite{ge2018multi}, ML-GCN~\cite{chen2019multi} and CMA~\cite{you2020cross}. The results on MS-COCO are reported in Table~\ref{tab:coco_result}.
The numbers of the compared methods are taken from the best reported results in their papers. 
It can be seen that our approach achieves the state-of-the-art results on all the metrics. It is also noticeable that for the second-best methods, i.e., ML-GCN and CMA, both of them leverage additional information of label correlations which is modeled by an extra graph neural network. In contrast, with the help of contrastive learning, our method achieves higher scores without using any additional information or neural networks.
\begin{table}%
\centering
\begin{tabular}{|c|c|c|c|}
\hline
Method & mAP & CF1 & OF1 \\
\hline\hline
CNN-RNN~\cite{wang2016cnn} & 61.2 & - & - \\
\hline
SRN~\cite{zhu2017learning} & 77.1  & 71.2 &  75.8 \\
\hline
s-CLs~\cite{liu2018multi} & 74.6 & 69.2 & 74.0\\
\hline
Multi Evidence~\cite{ge2018multi} & -  & 74.9  & 78.4\\
\hline
ML-GCN~\cite{chen2019multi} & 83.0 & \underline{78.0} & 80.3\\
\hline
CMA~\cite{you2020cross} & \underline{83.4} & 77.8 & \underline{80.9}\\
\hline\hline
MulCon (Ours) & \textbf{84.0} & \textbf{78.6} & \textbf{81.0} \\
\hline
\end{tabular}
\caption{Results on COCO dataset. The best and second scores are highlighted in boldface and with underlines, respectively.}
\label{tab:coco_result}
\end{table}

\subsection{Results on NUS-WIDE}
The NUS-WIDE dataset originally contained 269,648 images from Flicker and has been manually annotated with 81 visual concepts. Since some URLs have been deleted, we follow \cite{ben2020asymmetric} to obtain 200,000 images with 2.4 labels per image on average. We use the same hyperparameter setting as MS-COCO for training and testing on this dataset.
We also select the state-of-the-art methods that report the results on NUS-WIDE, including FitsNet~\cite{romero2014fitnets}, attention-transfer~\cite{zagoruyko2016paying}, s-CLs~\cite{liu2018multi}, CMA~\cite{you2020cross} and SRN~\cite{zhu2017learning}. 
The results on NUS-WIDE are shown in Table \ref{tab:nus_result}. We observe that MulCon performs the best on mAP and OF1.
Although MulCon is the second best on CF1, it outperforms CMA by 1.7\% on mAP, which is the most important metric in multi-label classification.
Moreover, SRN achieves good results on NUS-WIDE as well, however, its performance seems to be unstable as its results on MS-COCO are not comparable to ours.
\begin{table}%
\centering
\begin{tabular}{|c|c|c|c|}
\hline
Method & mAP & CF1 & OF1\\
\hline\hline
FitsNet~\cite{romero2014fitnets} & 57.4 & 54.9 & 70.4 \\
\hline
attention-transfer~\cite{zagoruyko2016paying} & 57.6 & 55.2 & 70.3 \\
\hline
s-CLs~\cite{liu2018multi} & 60.1 & 58.7 & 73.3  \\
\hline
CMA~\cite{you2020cross} & 60.8 & \textbf{60.4} & 73.7  \\
\hline
SRN~\cite{zhu2017learning} & \underline{62.0} & 58.5 & \underline{73.4} \\
\hline\hline
MulCon (Ours) & \textbf{62.5} & \underline{59.0} & \textbf{73.8}\\
\hline
\end{tabular}
\caption{Results on NUS-WIDE dataset. The best and second scores are highlighted in boldface and with underlines, respectively.}
\label{tab:nus_result}
\end{table}

\begin{table}
\centering
\begin{tabular}{|c|c|}
\hline
Method & mAP\\
\hline \hline
Backbone with BCE & 80.8\\
\hline
Backbone with BCE + SCL & 81.0 \\
\hline
MulCon with BCE only & 83.0\\
\hline
MulCon without pretraining & 81.1\\
\hline
MulCon & 84.0\\
\hline
\end{tabular}
\caption{Ablation study of different variants and training policies of MulCon.}
\label{tab:ablation}
\end{table}

\begin{figure*}[th]
\includegraphics[width=\linewidth]{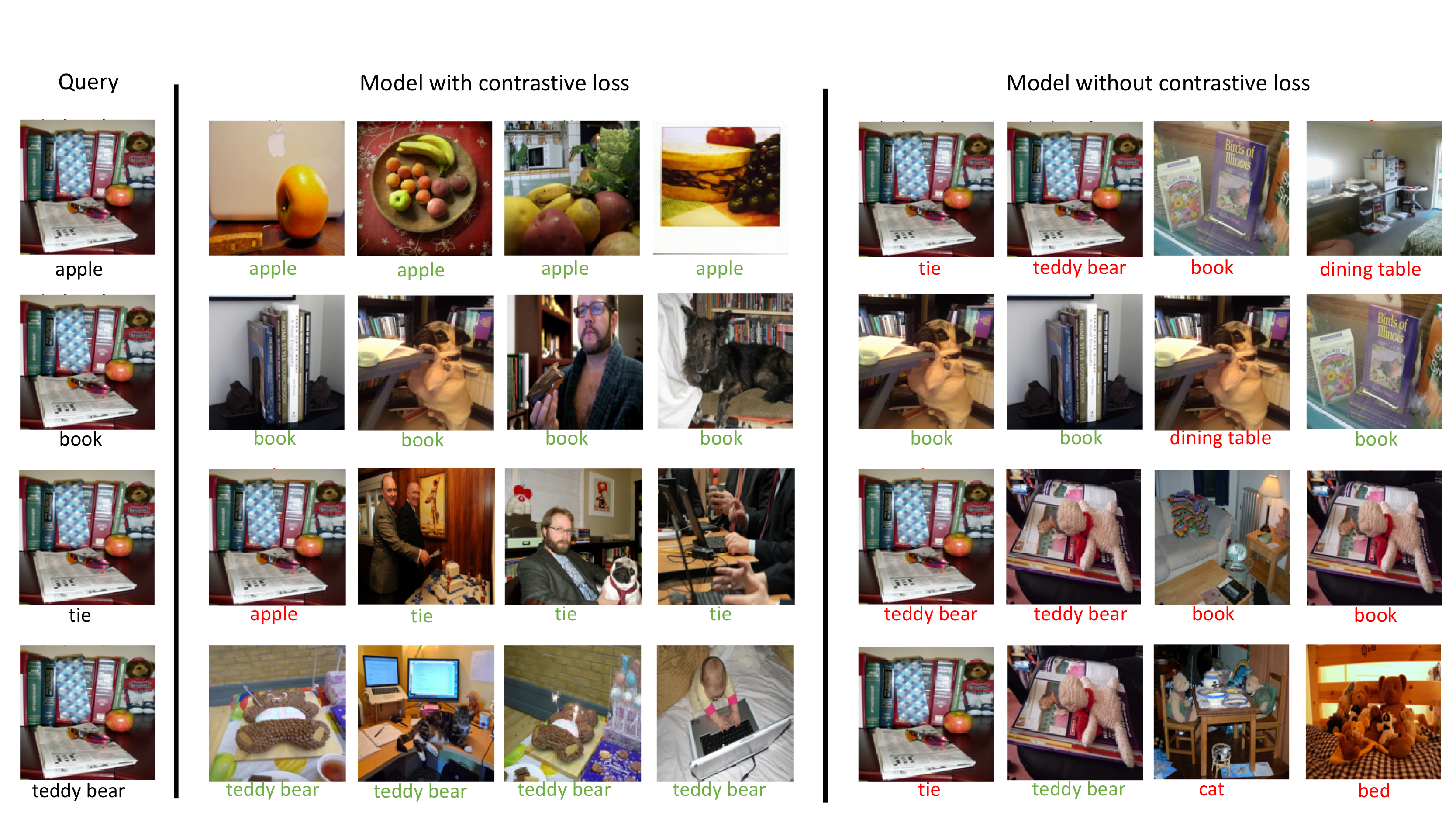}
   \caption{Top-4 related images retrieved given an query image and label on COCO dataset. 
   The results for our full model (MulCon) are on the left, and the results for our model without contrastive loss (MulCon with BCE only) are on the right. 
   The label under each retrieved image is the one corresponding to the embedding closest to the picked query embedding.}
\label{fig:retrieval_visual}
\vspace{-0.1in}
\end{figure*}
\begin{figure}
\includegraphics[width=\linewidth]{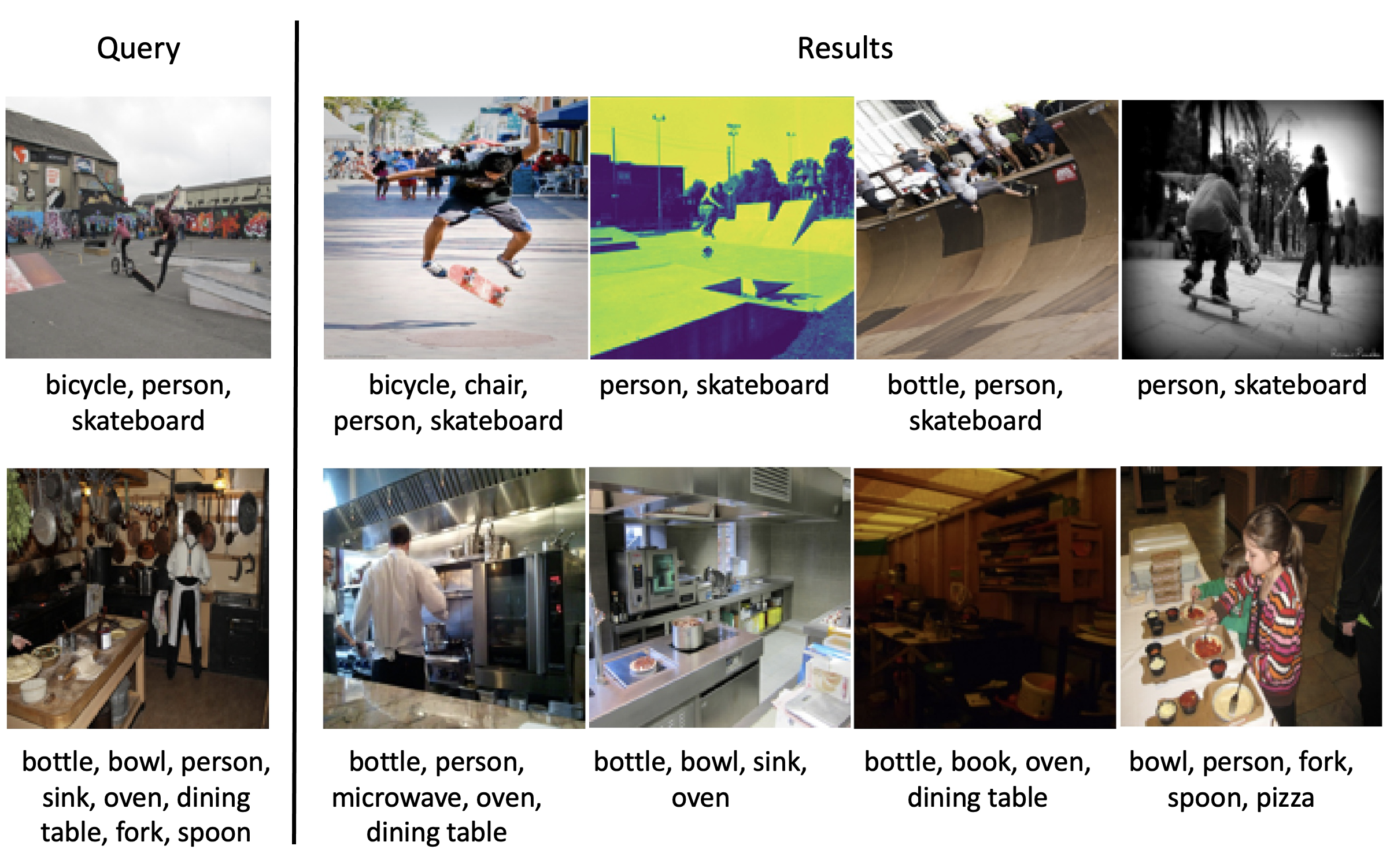}
   \caption{Top-4 related images retrieved given an query image and multiple labels on COCO dataset.}
\label{fig:retrieval_multi_label_visual}
\end{figure}
\vspace{-0.1in}
\begin{figure}
\centering
\includegraphics[width=\linewidth]{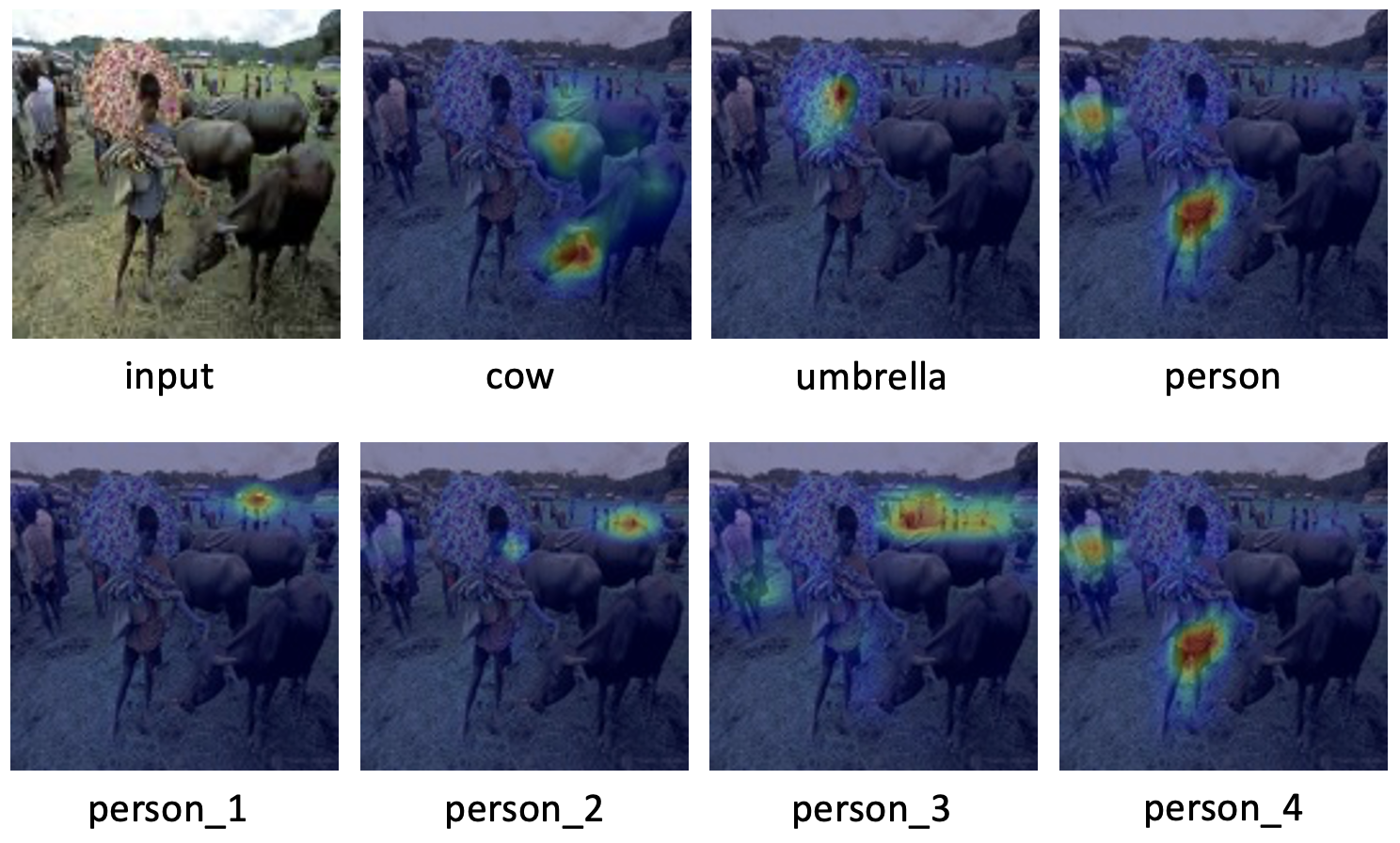}
   \caption{Visualization of attention maps. The top row includes the input image and the selected attention map for the ground-truth labels. The bottom row includes the multi-headed attention maps for class ``person''.}
\label{fig:attention_visual}
\vspace{-0.1in}
\end{figure}

\subsection{Ablation Study}
To fully understand the modules and training policies in MulCon,
we provide a comprehensive ablation study in this section.
Specifically, we are interested in the following variants of our method.

\begin{itemize}
    \item Backbone with BCE: Resnet101 trained with the BCE loss (i.e., Eq.~\ref{eq:bce}). This is a standard multi-label classification baseline which directly uses a classification network with image-level features.
    \item Backbone with BCE + SCL: Resnet101 trained with the BCE and  supervised-CL~\cite{khosla2020supervised} (SCL) losses (i.e., Eq.~\ref{eq:final_loss}). This is a variant where the supervised CL loss is directly applied on the image-level features.
    Note that for this variant, each image is an anchor, and thus we need to define the positive/negative sets for each anchor. Following the spirit of~\cite{khosla2020supervised}, we consider other images in the minibatch that have at least one common active label with the anchor as its positive samples and the others as the negative ones.
    \item MulCon with BCE only: MulCon without contrastive learning loss. This is the first pretraining step described in Section~\ref{sec:why_con_work}.
    \item MulCon without pretraining: This is the variant where we train MulCon with the BCE and SCL losses from the beginning, i.e., the second contrastive learning step described in Section~\ref{sec:why_con_work}.
    \item MulCon: The complete model of MulCon trained with the two-step policy described in Section~\ref{sec:why_con_work}.
\end{itemize}

Table \ref{tab:ablation} shows the results for the ablation study for our method in MS-COCO. 
We have the following remarks:
\textbf{1)} By comparing between Backbone with and without SCL, we can see that unlike in single-label classification, directly applying SCL on the image-level features cannot significantly improve the performance in multi-label cases (80.8 vs 81.0 mAP).
\textbf{2)} Note that the difference between Backbone with BCE and MulCon with BCE only lies in the label-level embedding network. It can be observed that the proposed label-level embedding network can significantly improve the mAP score even without the SCL loss (80.8 vs 83.0 mAP).
\textbf{3)} Here we compare among the different training policies of MulCon (i.e., the last three rows in Table~\ref{tab:ablation}). We observe that if the SCL loss is applied from the beginning of the training, it cannot improve the performance (83.of MulCon with BCE only vs 81.1 of MulCon without pretraining). This is because CL may over enforce distinctiveness which makes the model care less about label correlations, as analyzed in Section~\ref{sec:why_con_work}. However, if we pretrain MulCon with BCE and finetune with BCE+SCL (i.e., our complete model), it achieves better performance.

\subsection{Qualitative Analysis}

To qualitatively study our method, we first exam the semantics captured by the label-level embeddings learned with MulCon by conducting an image retrieval experiment. 
Specifically, given an input image $x_i$, we do a forward-pass to get its label-level embeddings $g_i$. With the ground-truth labels of $x_i$, we can specify a specific active label $j$, pick its embedding $g_{ij}$ from $g_i$ and use it as the query for image retrieval.
By comparing the Euclidean distance between the query embedding with the label-level embeddings of other images, we can retrieve the closest images to our query image and label. Figure~\ref{fig:retrieval_visual} shows the image retrieval results of MulCon and its variant with BCE only. It can be seen that %
with the help of contrastive learning, the retrieval results of MulCon are more accurate than its variant without CL.
To further demonstrate the intuitive meanings of the learned label-level embeddings in the multi-label setting, 
we provide an additional retrieval experiment where instead of using a single query label, we select multiple ones and concatenate their label-level embeddings as the query vector. That is to say, given an image, we can retrieve the images with multi-label labels of interest. Figure~\ref{fig:retrieval_multi_label_visual} shows the multi-label retrieval results.

We now show a visualization for the attention maps in terms of the active labels of an image produced by the multi-head attention used in the label-level embedding network. As illustrated in the top row of Figure \ref{fig:attention_visual}, the attention maps can precisely highlight the regions of the image in terms of each of its ground-truth labels.
In the bottom row of  Figure \ref{fig:attention_visual}, we show the attention maps of the heads of one label ``person''. It can be observed that the heads tend to capture the multiple instances of ``person'' in the image.

\section{Conclusion}

In this paper, we have introduced an end-to-end multi-label image classification framework, MulCon, which we believe is the first method that leverages contrastive learning in multi-label classification.
It has been shown that CL is not directly applicable in this domain, due to the fact that it is hard to define the positive/negative samples for an anchor with multiple labels.
To tackle this issue, we have introduced to learn label-level embeddings for an image with the multi-head attention mechanism. With label-level embeddings, we transform the multi-label classification into a single case for each label-level embedding, which facilities a straightforward adaption of supervised CL.
We have provided analytical and empirical study of why CL helps in multi-label learning and proposed a simple training policy to control the distinctiveness enforced by CL.
Extensive experimental results and visualization show the effectiveness of our approach and its ability to achieve the state-of-the-art performance in multi-label image classification.


\begin{thebibliography}{10}\itemsep=-1pt

\bibitem{ba2014multiple}
Jimmy Ba, Volodymyr Mnih, and Koray Kavukcuoglu.
\newblock Multiple object recognition with visual attention.
\newblock {\em arXiv preprint arXiv:1412.7755}, 2014.

\bibitem{bachman2019learning}
Philip Bachman, R~Devon Hjelm, and William Buchwalter.
\newblock Learning representations by maximizing mutual information across
  views.
\newblock {\em arXiv preprint arXiv:1906.00910}, 2019.

\bibitem{ben2020asymmetric}
Emanuel Ben-Baruch, Tal Ridnik, Nadav Zamir, Asaf Noy, Itamar Friedman, Matan
  Protter, and Lihi Zelnik-Manor.
\newblock Asymmetric loss for multi-label classification.
\newblock {\em arXiv preprint arXiv:2009.14119}, 2020.

\bibitem{chen2018order}
Shang-Fu Chen, Yi-Chen Chen, Chih-Kuan Yeh, and Yu-Chiang Wang.
\newblock Order-free rnn with visual attention for multi-label classification.
\newblock In {\em Proceedings of the AAAI Conference on Artificial
  Intelligence}, volume~32, 2018.

\bibitem{chen2020simple}
Ting Chen, Simon Kornblith, Mohammad Norouzi, and Geoffrey Hinton.
\newblock A simple framework for contrastive learning of visual
  representations.
\newblock In {\em International conference on machine learning}, pages
  1597--1607. PMLR, 2020.

\bibitem{chen2018recurrent}
Tianshui Chen, Zhouxia Wang, Guanbin Li, and Liang Lin.
\newblock Recurrent attentional reinforcement learning for multi-label image
  recognition.
\newblock In {\em Proceedings of the AAAI Conference on Artificial
  Intelligence}, volume~32, 2018.

\bibitem{chen2019learning}
Tianshui Chen, Muxin Xu, Xiaolu Hui, Hefeng Wu, and Liang Lin.
\newblock Learning semantic-specific graph representation for multi-label image
  recognition.
\newblock In {\em Proceedings of the IEEE/CVF International Conference on
  Computer Vision}, pages 522--531, 2019.

\bibitem{chen2019multi}
Zhao-Min Chen, Xiu-Shen Wei, Peng Wang, and Yanwen Guo.
\newblock Multi-label image recognition with graph convolutional networks.
\newblock In {\em Proceedings of the IEEE/CVF Conference on Computer Vision and
  Pattern Recognition}, pages 5177--5186, 2019.

\bibitem{chua2009nus}
Tat-Seng Chua, Jinhui Tang, Richang Hong, Haojie Li, Zhiping Luo, and Yantao
  Zheng.
\newblock Nus-wide: a real-world web image database from national university of
  singapore.
\newblock In {\em Proceedings of the ACM international conference on image and
  video retrieval}, pages 1--9, 2009.

\bibitem{cubuk2020randaugment}
Ekin~D Cubuk, Barret Zoph, Jonathon Shlens, and Quoc~V Le.
\newblock Randaugment: Practical automated data augmentation with a reduced
  search space.
\newblock In {\em Proceedings of the IEEE/CVF Conference on Computer Vision and
  Pattern Recognition Workshops}, pages 702--703, 2020.

\bibitem{deng2009imagenet}
Jia Deng, Wei Dong, Richard Socher, Li-Jia Li, Kai Li, and Li Fei-Fei.
\newblock Imagenet: A large-scale hierarchical image database.
\newblock In {\em 2009 IEEE conference on computer vision and pattern
  recognition}, pages 248--255. Ieee, 2009.

\bibitem{ge2018multi}
Weifeng Ge, Sibei Yang, and Yizhou Yu.
\newblock Multi-evidence filtering and fusion for multi-label classification,
  object detection and semantic segmentation based on weakly supervised
  learning.
\newblock In {\em Proceedings of the IEEE Conference on Computer Vision and
  Pattern Recognition}, pages 1277--1286, 2018.

\bibitem{ge2018chest}
Zongyuan Ge, Dwarikanath Mahapatra, Suman Sedai, Rahil Garnavi, and Rajib
  Chakravorty.
\newblock Chest x-rays classification: A multi-label and fine-grained problem.
\newblock {\em arXiv preprint arXiv:1807.07247}, 2018.

\bibitem{he2020momentum}
Kaiming He, Haoqi Fan, Yuxin Wu, Saining Xie, and Ross Girshick.
\newblock Momentum contrast for unsupervised visual representation learning.
\newblock In {\em Proceedings of the IEEE/CVF Conference on Computer Vision and
  Pattern Recognition}, pages 9729--9738, 2020.

\bibitem{he2016deep}
Kaiming He, Xiangyu Zhang, Shaoqing Ren, and Jian Sun.
\newblock Deep residual learning for image recognition.
\newblock In {\em Proceedings of the IEEE conference on computer vision and
  pattern recognition}, pages 770--778, 2016.

\bibitem{hjelm2018learning}
R~Devon Hjelm, Alex Fedorov, Samuel Lavoie-Marchildon, Karan Grewal, Phil
  Bachman, Adam Trischler, and Yoshua Bengio.
\newblock Learning deep representations by mutual information estimation and
  maximization.
\newblock {\em arXiv preprint arXiv:1808.06670}, 2018.

\bibitem{huynh2020interactive}
Dat Huynh and Ehsan Elhamifar.
\newblock Interactive multi-label cnn learning with partial labels.
\newblock In {\em Proceedings of the IEEE/CVF Conference on Computer Vision and
  Pattern Recognition}, pages 9423--9432, 2020.

\bibitem{huynh2020shared}
Dat Huynh and Ehsan Elhamifar.
\newblock A shared multi-attention framework for multi-label zero-shot
  learning.
\newblock In {\em Proceedings of the IEEE/CVF Conference on Computer Vision and
  Pattern Recognition}, pages 8776--8786, 2020.

\bibitem{jain2016extreme}
Himanshu Jain, Yashoteja Prabhu, and Manik Varma.
\newblock Extreme multi-label loss functions for recommendation, tagging,
  ranking \& other missing label applications.
\newblock In {\em Proceedings of the 22nd ACM SIGKDD International Conference
  on Knowledge Discovery and Data Mining}, pages 935--944, 2016.

\bibitem{khosla2020supervised}
Prannay Khosla, Piotr Teterwak, Chen Wang, Aaron Sarna, Yonglong Tian, Phillip
  Isola, Aaron Maschinot, Ce Liu, and Dilip Krishnan.
\newblock Supervised contrastive learning.
\newblock {\em arXiv preprint arXiv:2004.11362}, 2020.

\bibitem{kim2020adversarial}
Minseon Kim, Jihoon Tack, and Sung~Ju Hwang.
\newblock Adversarial self-supervised contrastive learning.
\newblock {\em arXiv preprint arXiv:2006.07589}, 2020.

\bibitem{kingma2014adam}
Diederik~P Kingma and Jimmy Ba.
\newblock Adam: A method for stochastic optimization.
\newblock {\em arXiv preprint arXiv:1412.6980}, 2014.

\bibitem{lee2019set}
Juho Lee, Yoonho Lee, Jungtaek Kim, Adam Kosiorek, Seungjin Choi, and Yee~Whye
  Teh.
\newblock Set transformer: A framework for attention-based
  permutation-invariant neural networks.
\newblock In {\em International Conference on Machine Learning}, pages
  3744--3753. PMLR, 2019.

\bibitem{li2020prototypical}
Junnan Li, Pan Zhou, Caiming Xiong, Richard Socher, and Steven~CH Hoi.
\newblock Prototypical contrastive learning of unsupervised representations.
\newblock {\em arXiv preprint arXiv:2005.04966}, 2020.

\bibitem{li2016conditional}
Qiang Li, Maoying Qiao, Wei Bian, and Dacheng Tao.
\newblock Conditional graphical lasso for multi-label image classification.
\newblock In {\em Proceedings of the IEEE Conference on Computer Vision and
  Pattern Recognition}, pages 2977--2986, 2016.

\bibitem{li2014multi}
Xin Li, Feipeng Zhao, and Yuhong Guo.
\newblock Multi-label image classification with a probabilistic label
  enhancement model.
\newblock In {\em UAI}, volume~1, pages 1--10, 2014.

\bibitem{li2016human}
Yining Li, Chen Huang, Chen~Change Loy, and Xiaoou Tang.
\newblock Human attribute recognition by deep hierarchical contexts.
\newblock In {\em European Conference on Computer Vision}, pages 684--700.
  Springer, 2016.

\bibitem{lin2017focal}
Tsung-Yi Lin, Priya Goyal, Ross Girshick, Kaiming He, and Piotr Doll{\'a}r.
\newblock Focal loss for dense object detection.
\newblock In {\em Proceedings of the IEEE international conference on computer
  vision}, pages 2980--2988, 2017.

\bibitem{lin2014microsoft}
Tsung-Yi Lin, Michael Maire, Serge Belongie, James Hays, Pietro Perona, Deva
  Ramanan, Piotr Doll{\'a}r, and C~Lawrence Zitnick.
\newblock Microsoft coco: Common objects in context.
\newblock In {\em European conference on computer vision}, pages 740--755.
  Springer, 2014.

\bibitem{liu2017semantic}
Feng Liu, Tao Xiang, Timothy~M Hospedales, Wankou Yang, and Changyin Sun.
\newblock Semantic regularisation for recurrent image annotation.
\newblock In {\em Proceedings of the IEEE Conference on Computer Vision and
  Pattern Recognition}, pages 2872--2880, 2017.

\bibitem{liu2018multi}
Yongcheng Liu, Lu Sheng, Jing Shao, Junjie Yan, Shiming Xiang, and Chunhong
  Pan.
\newblock Multi-label image classification via knowledge distillation from
  weakly-supervised detection.
\newblock In {\em Proceedings of the 26th ACM international conference on
  Multimedia}, pages 700--708, 2018.

\bibitem{luong2015effective}
Minh-Thang Luong, Hieu Pham, and Christopher~D Manning.
\newblock Effective approaches to attention-based neural machine translation.
\newblock {\em arXiv preprint arXiv:1508.04025}, 2015.

\bibitem{medsker2001recurrent}
Larry~R Medsker and LC Jain.
\newblock Recurrent neural networks.
\newblock {\em Design and Applications}, 5, 2001.

\bibitem{oord2018representation}
Aaron van~den Oord, Yazhe Li, and Oriol Vinyals.
\newblock Representation learning with contrastive predictive coding.
\newblock {\em arXiv preprint arXiv:1807.03748}, 2018.

\bibitem{radford2021learning}
Alec Radford, Jong~Wook Kim, Chris Hallacy, Aditya Ramesh, Gabriel Goh,
  Sandhini Agarwal, Girish Sastry, Amanda Askell, Pamela Mishkin, Jack Clark,
  et~al.
\newblock Learning transferable visual models from natural language
  supervision.
\newblock {\em arXiv preprint arXiv:2103.00020}, 2021.

\bibitem{romero2014fitnets}
Adriana Romero, Nicolas Ballas, Samira~Ebrahimi Kahou, Antoine Chassang, Carlo
  Gatta, and Yoshua Bengio.
\newblock Fitnets: Hints for thin deep nets.
\newblock {\em arXiv preprint arXiv:1412.6550}, 2014.

\bibitem{van2008visualizing}
Laurens Van~der Maaten and Geoffrey Hinton.
\newblock Visualizing data using t-sne.
\newblock {\em Journal of machine learning research}, 9(11), 2008.

\bibitem{vaswani2017attention}
Ashish Vaswani, Noam Shazeer, Niki Parmar, Jakob Uszkoreit, Llion Jones,
  Aidan~N Gomez, Lukasz Kaiser, and Illia Polosukhin.
\newblock Attention is all you need.
\newblock {\em arXiv preprint arXiv:1706.03762}, 2017.

\bibitem{wang2016cnn}
Jiang Wang, Yi Yang, Junhua Mao, Zhiheng Huang, Chang Huang, and Wei Xu.
\newblock Cnn-rnn: A unified framework for multi-label image classification.
\newblock In {\em Proceedings of the IEEE conference on computer vision and
  pattern recognition}, pages 2285--2294, 2016.

\bibitem{wang2021exploring}
Wenguan Wang, Tianfei Zhou, Fisher Yu, Jifeng Dai, Ender Konukoglu, and Luc
  Van~Gool.
\newblock Exploring cross-image pixel contrast for semantic segmentation.
\newblock {\em arXiv preprint arXiv:2101.11939}, 2021.

\bibitem{wang2017multi}
Zhouxia Wang, Tianshui Chen, Guanbin Li, Ruijia Xu, and Liang Lin.
\newblock Multi-label image recognition by recurrently discovering attentional
  regions.
\newblock In {\em Proceedings of the IEEE international conference on computer
  vision}, pages 464--472, 2017.

\bibitem{xu2015show}
Kelvin Xu, Jimmy Ba, Ryan Kiros, Kyunghyun Cho, Aaron Courville, Ruslan
  Salakhudinov, Rich Zemel, and Yoshua Bengio.
\newblock Show, attend and tell: Neural image caption generation with visual
  attention.
\newblock In {\em International conference on machine learning}, pages
  2048--2057. PMLR, 2015.

\bibitem{yang2016improving}
Hao Yang, Joey~Tianyi Zhou, and Jianfei Cai.
\newblock Improving multi-label learning with missing labels by structured
  semantic correlations.
\newblock In {\em European conference on computer vision}, pages 835--851.
  Springer, 2016.

\bibitem{yang2015pinterest}
Xitong Yang, Yuncheng Li, and Jiebo Luo.
\newblock Pinterest board recommendation for twitter users.
\newblock In {\em Proceedings of the 23rd ACM international conference on
  Multimedia}, pages 963--966, 2015.

\bibitem{ye2020attention}
Jin Ye, Junjun He, Xiaojiang Peng, Wenhao Wu, and Yu Qiao.
\newblock Attention-driven dynamic graph convolutional network for multi-label
  image recognition.
\newblock In {\em European Conference on Computer Vision}, pages 649--665.
  Springer, 2020.

\bibitem{you2020cross}
Renchun You, Zhiyao Guo, Lei Cui, Xiang Long, Yingze Bao, and Shilei Wen.
\newblock Cross-modality attention with semantic graph embedding for
  multi-label classification.
\newblock In {\em Proceedings of the AAAI Conference on Artificial
  Intelligence}, volume~34, pages 12709--12716, 2020.

\bibitem{zagoruyko2016paying}
Sergey Zagoruyko and Nikos Komodakis.
\newblock Paying more attention to attention: Improving the performance of
  convolutional neural networks via attention transfer.
\newblock {\em arXiv preprint arXiv:1612.03928}, 2016.

\bibitem{zhou2018graph}
Jie Zhou, Ganqu Cui, Zhengyan Zhang, Cheng Yang, Zhiyuan Liu, Lifeng Wang,
  Changcheng Li, and Maosong Sun.
\newblock Graph neural networks: A review of methods and applications.
\newblock {\em arXiv preprint arXiv:1812.08434}, 2018.

\bibitem{zhu2017learning}
Feng Zhu, Hongsheng Li, Wanli Ouyang, Nenghai Yu, and Xiaogang Wang.
\newblock Learning spatial regularization with image-level supervisions for
  multi-label image classification.
\newblock In {\em Proceedings of the IEEE Conference on Computer Vision and
  Pattern Recognition}, pages 5513--5522, 2017.

\end{thebibliography}
\end{document}